\documentclass{article}
\usepackage{iclr2027_conference,times}
\usepackage[T1]{fontenc}
\usepackage{amsmath,amsfonts,bm}

\def\eqref#1{equation~\ref{#1}}
\def\1{\bm{1}}

\DeclareMathAlphabet{\mathsfit}{\encodingdefault}{\sfdefault}{m}{sl}
\SetMathAlphabet{\mathsfit}{bold}{\encodingdefault}{\sfdefault}{bx}{n}

\usepackage{amsmath,amssymb}
\usepackage{booktabs,array}
\usepackage{graphicx}
\newsavebox{\papertablebox}
\usepackage{xcolor}
\usepackage{colortbl}
\definecolor{ablDarkRed}{HTML}{8B0000}
\definecolor{ablDarkGreen}{HTML}{006400}
\newlength{\ablDigitWidth}
\newlength{\ablPointWidth}
\newlength{\ablSignWidth}
\newlength{\ablParenWidth}
\newlength{\ablMarkWidth}
\newlength{\ablIntegerWidth}
\newlength{\ablDeltaWidth}
\makeatletter
\newcommand{\ablDigits}[1]{%
  \@tfor\ablCurrentDigit:=#1\do{\makebox[\ablDigitWidth][c]{\ablCurrentDigit}}%
}
\makeatother
\newcommand{\ablPlain}[1]{#1}
\def\ablNumberParts#1.#2\ablStop{%
  \ablDigits{#1}\makebox[\ablPointWidth][c]{.}\ablDigits{#2}%
}
\newcommand{\ablNumber}[2]{%
  \makebox[\dimexpr\ablIntegerWidth+\ablPointWidth+\ablDigitWidth\relax][r]{#1{\ablNumberParts#2\ablStop}}%
}
\newcommand{\ablDelta}[3]{%
  \textcolor{#1}{%
    \makebox[\ablDeltaWidth][l]{%
      \makebox[\ablParenWidth][c]{(}%
      \ensuremath{#2}\ablNumberParts#3\ablStop%
      \makebox[\ablParenWidth][c]{)}%
    }%
  }%
}
\newcommand{\ablSelf}{%
  \makebox[\ablParenWidth][c]{(}%
  \textemdash%
  \makebox[\ablParenWidth][c]{)}%
}
\newcommand{\ablCell}[4]{%
  \ablNumber{#1}{#2}%
  \makebox[\ablMarkWidth][l]{#3}%
  \hspace{0.15em}%
  \makebox[\ablDeltaWidth][l]{#4}%
}

\newcommand{\ablSetup}{%
  \setlength{\tabcolsep}{3pt}%
  \renewcommand{\arraystretch}{1.12}%
  \settowidth{\ablDigitWidth}{\textbf{0}}%
  \settowidth{\ablPointWidth}{.}%
  \settowidth{\ablSignWidth}{$+$}%
  \settowidth{\ablParenWidth}{(}%
  \settowidth{\ablMarkWidth}{$^{\dagger}$}%
  \setlength{\ablIntegerWidth}{3\ablDigitWidth}%
  \setlength{\ablDeltaWidth}{\dimexpr2\ablParenWidth+\ablSignWidth+3\ablDigitWidth+\ablPointWidth\relax}%
}

\newcommand{\ablUseTable}{%
  \ifdim\wd\papertablebox>\linewidth%
  \resizebox{\linewidth}{!}{\usebox{\papertablebox}}%
  \else%
  \usebox{\papertablebox}%
  \fi%
}

\definecolor{green}{RGB}{0,128,0}
\usepackage{microtype}
\usepackage{placeins}
\usepackage{hyperref}
\usepackage{url}
\title{Evo-World: A Lightweight Vision-Language-Action Model with Future Motion Prediction via Optical-Flow Supervision}
\author{%
{\bf Zile Zhong}$^{1}$\thanks{Equal contribution.}\quad
{\bf Yuxin Du}$^{1}$\footnotemark[1]\quad
{\bf Tao Lin}$^{1}$\footnotemark[1]\quad
{\bf Haoxiao Wang}$^{2}$\quad
{\bf Wei Dai}$^{3}$\quad
{\bf Binhua Zuo}$^{4}$\quad
{\bf Bo Zhao}$^{1}$\thanks{Corresponding author.}\\[5pt]
$^{1}$School of AI, Shanghai Jiao Tong University\quad
$^{2}$Zhejiang University\\
$^{3}$University of Electronic Science and Technology of China\quad
$^{4}$QuicRobot\\[5pt]
\texttt{zilezhong2005@gmail.com}\quad\texttt{bo.zhao@sjtu.edu.cn}\\[5pt]
\url{https://github.com/MINT-SJTU/Evo-World}%
}

\iclrfinalcopy

\makeatletter
\def\@maketitle{%
  \vbox{%
    \hsize\textwidth
    {\LARGE\sc \@title\par}%
    \vskip 0.25in%
    {\normalsize \@author\par}%
    \vskip 0.2in%
  }%
}
\makeatother
\hypersetup{
  pdftitle={Evo-World: A Lightweight Vision-Language-Action Model with Future Motion Prediction via Optical-Flow Supervision},
  pdfauthor={Zile Zhong, Yuxin Du, Tao Lin, Haoxiao Wang, Wei Dai, Bo Zhao}
}

\begin{document}
\maketitle
\begin{abstract}
% Generated from content/1_Abstract.md by review/sync_tex.py. Edit the Markdown source.

Vision-language-action (VLA) models provide a unified framework for visual understanding, language grounding, and robot control, yet anticipating how the scene will evolve beyond the current observation can provide additional information for control. Action supervision alone does not explicitly encourage such anticipation, while future-image prediction entangles task-relevant motion with largely static appearance. To address these limitations, we introduce \textbf{Evo-World}, a lightweight VLA model that augments policy learning with future motion prediction through optical-flow supervision, thereby providing a task-relevant predictive signal focused on motion rather than appearance. We further design the training procedure to preserve these predictive representations during policy learning, regulating representation diversity while decoupling representation learning from action optimization. Across simulation benchmarks, Evo-World demonstrates strong performance, achieving a tier-averaged success rate of \textbf{91.1\%} on Meta-World MT50 , with a 13.0\% relative improvement over the non-predictive baseline. On real-world AgileX robots, Evo-World outperforms the evaluated baselines across single-arm and bimanual manipulation tasks while using only \textbf{0.87B} parameters and delivering lower GPU memory usage and higher inference efficiency. These results demonstrate that future motion prediction through optical-flow supervision provides an effective and lightweight approach to enhancing the predictive capabilities of VLA models. 

\end{abstract}

% Maintained directly in this standalone project; aligned with the current abstract.

\section{Introduction}

Vision-language-action (VLA) models provide a unified framework for visual understanding, language grounding, and robot control. By adapting pretrained visual and linguistic knowledge to action generation, they support a broad range of manipulation behaviors \citep{RT2,OpenVLA,Pi0}. Successful manipulation also depends on anticipating how a scene should change as a task unfolds: grasping, placing, and opening require coordinated changes in the robot and its surroundings. Yet action supervision alone does not explicitly require a policy's representations to anticipate these changes. Providing this predictive structure is therefore an opportunity to improve VLA policies beyond their understanding of the current observation.

\begin{figure}[htbp]
\centering
\includegraphics[width=\linewidth]{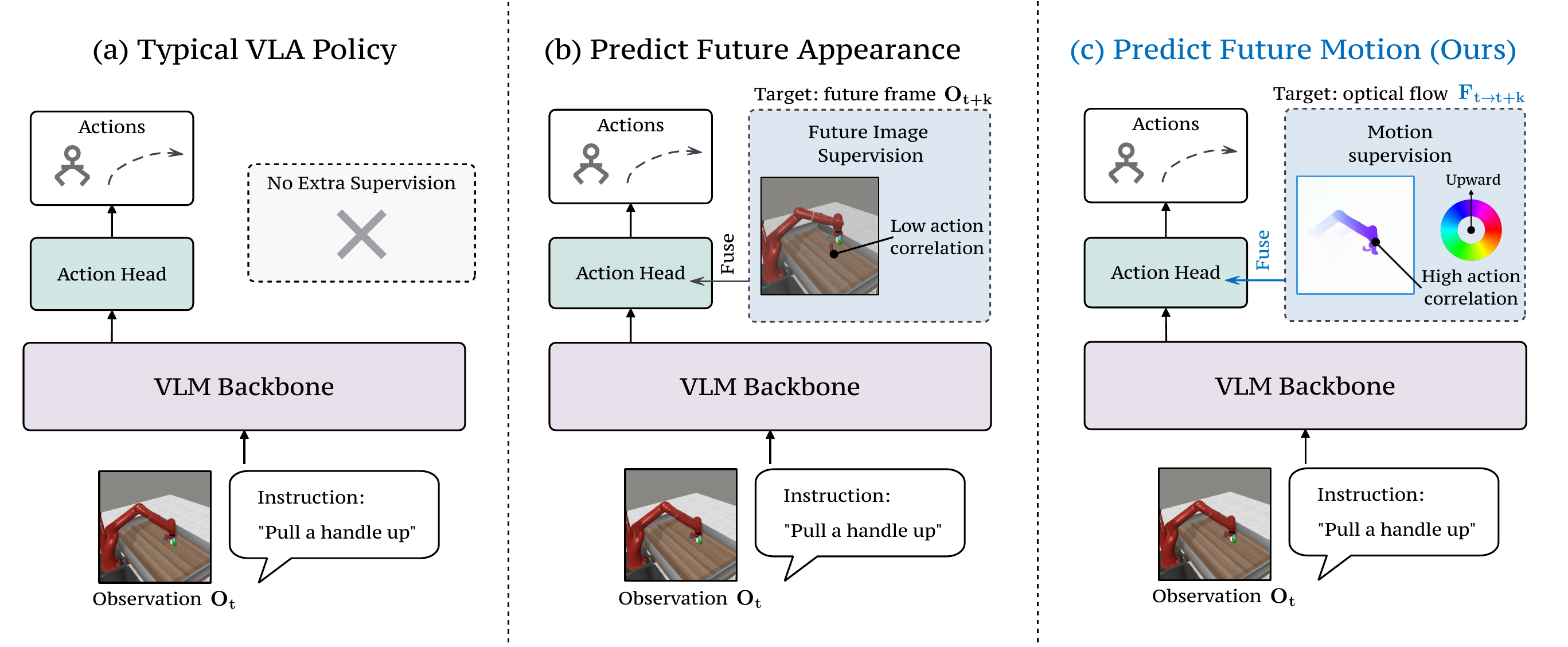}
\caption{\textbf{Prediction targets for a VLA policy.} (a) An action-only policy; (b) prediction of future appearance; and (c) prediction of future motion with optical-flow supervision. The two predictive branches condition action generation on latent representations. The flow illustration depicts the training target. The action-correlation labels express the motivation for motion supervision qualitatively.}
\label{fig:1}
\end{figure}

Future-observation prediction offers temporal supervision for learning such anticipation \citep{GR1,GR2}, but the prediction target determines which aspects of scene evolution the model must represent. Future RGB images entangle motion with appearance, including background textures and objects that remain largely unchanged. Preserving these details can support visual recognition, but it does not directly emphasize the changes needed for control. Optical flow provides a complementary target by describing image-space displacement while suppressing static texture. This motivates learning future motion representations that complement the policy's visual and linguistic context, as illustrated in Figure~\ref{fig:1}.

We introduce \textbf{Evo-World}, a lightweight VLA model that augments policy learning with future motion prediction through optical-flow supervision. Evo-World learns to anticipate motion from the current observation and task instruction, and uses this prediction to inform action generation. The key idea is to make scene change an effective learning target while retaining the appearance and semantics available to the vision-language backbone. Optical flow computed from demonstration videos supplies supervision during training; a compact predictive branch learns its latent representation. 

For the predictive signal to remain useful, policy learning must preserve informative motion representations. We address two complementary requirements: maintaining representation diversity and preserving the motion-learning objective during action optimization. We adopt Sketched Isotropic Gaussian Regularization (SIGReg) from LeJEPA to discourage collapse of learned representations \citep{LeJEPA}. We also separate the gradient paths for representation learning and action optimization, allowing the policy to learn how to use the motion features while the predictive branch is trained by its own objective. 

Evo-World contains only \textbf{0.87B} parameters, including its predictive branch. At deployment, it uses current camera observations, the instruction, and robot state; future observations and optical-flow estimation are unnecessary. Motion prediction operates in a latent space, avoiding the need to decode future images or flow fields during action generation. This design incorporates anticipation within a compact policy suitable for deployment on a physical robot.

We evaluate Evo-World on Meta-World MT50, RoboTwin, and three real-world manipulation tasks~\citep{Meta-World}\citep{RoboTwin2}. It achieves \textbf{91.1\%} tier-averaged success on Meta-World, a 13.0\% relative improvement over the non-predictive baseline, and \textbf{41.8\% }average success across Clean2Clean and Clean2Random on RoboTwin (Table~\ref{tab:1}). On an AgileX dual-arm robot, Evo-World outperforms OpenVLA-OFT and $\pi_{0}$ on all three tasks, with 63.3\% average success (Table~\ref{tab:2}). These tasks cover single-arm floral arrangement and bimanual collaboration in mug hanging and microwave placement. Component ablations support the training design: regularization counters representation contraction, and both regularization and gradient separation improve aggregate success in the tested setting.

Our contributions are threefold:
\vspace{-5pt}

\begin{enumerate}
\item We introduce optical-flow-supervised future motion prediction into a lightweight VLA policy, providing a strong predictive signal that complements visual understanding and language grounding.
\item We combine distribution regularization with separate gradient paths to preserve predictive representations during policy learning, and integrate these representations into action generation through a compact conditioning module.
\item We conduct comprehensive experiments across simulation and real-world robotic manipulation tasks. With a compact 0.87B-parameter model, Evo-World achieves strong task performance while retaining low computational overhead.
\end{enumerate}

% Maintained directly in this standalone project; aligned with the current abstract.

\section{Related Work}

\subsection{Vision-language-action policies and efficient deployment}

VLA models connect visual understanding and language grounding to robot control. RT-2 and OpenVLA adapt pretrained vision-language models by predicting action tokens \citep{RT2,OpenVLA}. For continuous control, Diffusion Policy learns action sequences through denoising \citep{DiffusionPolicy} and $\pi_0$ integrates a flow-matching action expert with a vision-language backbone \citep{Pi0}. SmolVLA and Evo-1 emphasize compact architectures and efficient adaptation of multimodal features \citep{SmolVLA,Evo1}. These advances improve policy capacity and deployment practicality, but action supervision does not explicitly require anticipation of future scene changes.

\subsection{Predictive learning for robot control}

Predictive learning supplements action labels with supervision about how observations evolve. GR-1 and GR-2 learn future RGB observations alongside actions \citep{GR1,GR2}. V-JEPA 2 learns video embeddings through masked prediction and action-conditioned latent dynamics for planning \citep{VJEPA2}. Moto learns discrete motion tokens through conditional frame reconstruction and transfers motion priors to control \citep{Moto}. These methods connect visual prediction to control through both generative and latent representations. The prediction target determines which scene properties the model must preserve: RGB reconstruction couples motion with appearance, while learned representations inherit the structure of their supervision. Motion-specific targets offer a complementary way to emphasize displacement.

\subsection{Motion supervision for manipulation}

Motion guides manipulation through point trajectories \citep{ATM,Track2Act}, predicted flow \citep{Im2Flow2Act,RoboFlow4D}, and motion-guided visual prediction \citep{FLIP,FlowDreamer,FlowVLA,F2FAP}. Recent flow-supervised policies use this signal in different roles. \citet{MotionImageDiffusion} augment VLA training with motion-image diffusion in a frozen VAE latent space. This auxiliary head supervises the shared VLM and is omitted from policy inference. FlowWAM jointly models RGB and flow latents, using generated flow for action prediction and supplied flow to condition world modeling \citep{FlowWAM}. LaMP supplies future 3D motion priors through a frozen expert's one-step denoising features, avoiding full motion reconstruction \citep{LaMP}. TemporalFlow-VLA instead supervises execution-history queries with robot-surface flow derived from kinematics and camera calibration \citep{TemporalFlowVLA}.

Evo-World predicts future motion by directly regressing learned embeddings of 2D optical-flow renderings from the current observation and instruction. These embeddings remain available to action generation without motion-denoising or flow-reconstruction objectives. The motion branch continues optimizing its prediction objective and distribution regularization \citep{LeJEPA}, with action gradients blocked; bounded FiLM combines current and predicted embeddings to condition the policy \citep{FiLM}. Its RGB-derived targets require neither depth nor robot-geometry labels.

% Generated from content/4_Method.md by review/sync_tex.py. Edit the Markdown source.

\section{Method}
\label{sec:method}
\begin{figure}[htbp]
\centering
\includegraphics[width=\linewidth]{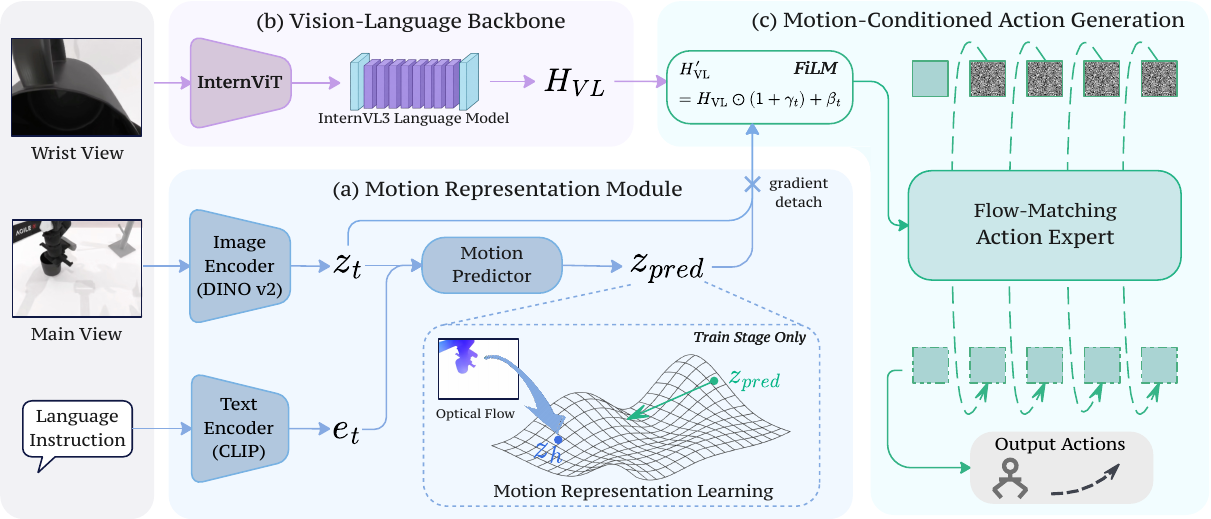}
\caption{\textbf{Evo-World architecture.} (a) The Motion Representation Module predicts a latent motion representation from DINOv2 image features and CLIP instruction features. In the dashed box, DINOv2 encodes an optical-flow RGB rendering as a target used only during training. (b) The InternVL3-1B vision-language backbone provides visual and linguistic context. (c) The current and predicted representations are concatenated and detached from action gradients, then projected into FiLM scale and shift parameters. These parameters modulate the backbone features to condition the flow-matching action expert for action generation.}
\label{fig:2}
\end{figure}

Given camera observations $O_t$, a language instruction $\ell$, and robot state $s_t$, Evo-World predicts an action chunk $a_{t:t+H-1}$. Figure~\ref{fig:2} gives an overview of the framework, which we describe in the following four parts.

\subsection{Motion Representation Module}

This world-model branch predicts latent motion from one designated view $O_t^w$ and the instruction. Encoders $E_I$ and $E_T$, initialized from DINOv2 and CLIP, respectively, followed by projection heads $P_I$ and $P_T$, produce $D$-dimensional features $z_t=P_I(E_I(O_t^w))$ and $e_\ell=P_T(E_T(\ell))$ \citep{DINOv2,CLIP}. A multilayer perceptron predicts instruction-conditioned demonstration motion over an interval $\Delta$ as $z_{\mathrm{pred}}=P_M([z_t;e_\ell])$. The pair $(z_t,z_{\mathrm{pred}})$ represents the current scene and anticipated motion, without candidate-action conditioning or flow decoding.

During training, the prediction target is encoded from a three-channel optical-flow rendering. We precompute $F_{t\rightarrow t+\Delta}=\operatorname{RAFT}(O_t^w,O_{t+\Delta}^w)$ with $\Delta=15$ \citep{RAFT}. A renderer $\mathcal R$ with a fixed magnitude scale produces $Y_t=\mathcal R(F_{t\rightarrow t+\Delta})$, which is encoded as $z_h=P_I(E_I(Y_t))$. The current and target images share $E_I,P_I$. The optical-flow processing results for RoboTwin and Meta-World are presented in Figure~\ref{fig:3}. The target pathway is used only during training.

\subsection{Vision-Language Backbone}

The available camera views are first encoded into visual tokens by the InternViT vision encoder. These visual tokens are then combined with the instruction $\ell$ and processed by the InternVL3 LLM to obtain the joint vision-language representation $H_{\mathrm{VL}} = E_{\mathrm{LLM}}\!\left(E_{\mathrm{ViT}}(O_t), \ell\right),$
where $E_{\mathrm{ViT}}$ denotes the InternViT encoder and $E_{\mathrm{LLM}}$ denotes the InternVL3 LLM. The resulting tokens provide visual and linguistic context for action generation \citep{Evo1,InternVL3}.

\subsection{Motion-Conditioned Action Generation}

FiLM computes scale and shift from the current and predicted representations as $(\gamma_t,\beta_t)=\alpha\tanh\!\left(G_\phi([\operatorname{sg}(z_t);\operatorname{sg}(z_{\mathrm{pred}})])\right)$ and modulates the vision-language tokens \citep{FiLM}:

\begin{equation}
H'_{\mathrm{VL}}
=H_{\mathrm{VL}}\odot(1+\gamma_t)+\beta_t.
\tag{1}
\label{eq:method-1}
\end{equation}

Modulation is bounded by $\alpha=0.1$ and broadcast across tokens. The generator's final layer is zero-initialized, giving identity modulation. Stop-gradient $\operatorname{sg}$ blocks action gradients into the motion representation module while allowing the action loss to train FiLM and the policy.

The action expert $A_\theta$ is a flow-matching transformer \citep{Evo1,FlowMatching}. It embeds noisy actions as tokens with temporal positions, cross-attends to $H'_{\mathrm{VL}}$ and an MLP-encoded state token, and injects a time embedding into each feed-forward block. An output MLP maps the aggregated action tokens to chunk velocities.

For a normalized action chunk $a$, draw $u\sim\mathcal U(-1,1)$ and set $\widetilde u=M\odot u$, where the binary mask $M$ excludes padded action coordinates. During training, we sample $\tau\sim\operatorname{Beta}(2,2)$, clip it to $[0.02,0.98]$, and interpolate $x_\tau=(1-\tau)\widetilde u+\tau a$. The expert predicts the velocity $\hat v_\theta=A_\theta(x_\tau,\tau,s_t,H'_{\mathrm{VL}})$.

At inference, we start from $x^0=\widetilde u$ and apply $N$ forward-Euler steps:

\begin{equation}
x^{k+1}=M\odot\left[x^k+\frac{1}{N}
A_\theta(x^k,k/N,s_t,H'_{\mathrm{VL}})\right].
\tag{2}
\label{eq:method-2}
\end{equation}

Here, $k=0,\ldots,N-1$, and the resulting chunk is $\hat a=x^N$. The motion representation module remains active; no future image or optical-flow estimator is required.

\subsection{Training Objectives}

\textbf{World loss.} For a batch of $B$ samples, we jointly optimize a prediction loss and a regularization loss as world loss. The prediction loss measures the discrepancy between the predicted and target motion representations, with gradients retained on both sides. To discourage representation collapse, we further apply Sketched Isotropic Gaussian Regularization (SIGReg) to both the current and target batch representations $z_t$ and $z_h$ \citep{LeJEPA}:
\begin{equation}
\mathcal L_{\mathrm{pred}}
=\frac{1}{BD}\sum_{b=1}^{B}
\|z_{\mathrm{pred},b}-z_{h,b}\|_2^2,
\qquad
\mathcal L_{\mathrm{SIGReg}}
=\frac{1}{2}\left[\mathcal S(z_t)+\mathcal S(z_h)\right].
\tag{3}
\label{eq:method-3}
\end{equation}

The statistic $\mathcal S$ matches characteristic functions of random one-dimensional projections to a standard Gaussian; its discrete form is given in the Appendix~\ref{app:sigreg_sta}. SIGReg directly regularizes $z_t,z_h$, influencing $z_{\mathrm{pred}}$ through the prediction loss.

\textbf{Action loss.} With target velocity $v^*=a-\widetilde u$, we average the squared velocity error over valid coordinates $j$, including the chunk dimension:

\begin{equation}
\mathcal L_{\mathrm{action}}
=\frac{\sum_{b,j}M_{b,j}(\hat v_{\theta,b,j}-v^*_{b,j})^2}
{\sum_{b,j}M_{b,j}+10^{-8}}.
\tag{4}
\label{eq:method-4}
\end{equation}

\textbf{Joint objective.} We optimize

\begin{equation}
\mathcal L
=\mathcal L_{\mathrm{action}}
+\lambda_{\mathrm{world}}
\left(\mathcal L_{\mathrm{pred}}
+\lambda_{\mathrm{sig}}\mathcal L_{\mathrm{SIGReg}}\right).
\tag{5}
\label{eq:method-5}
\end{equation}
% Maintained directly in this standalone project; no local Markdown source.

\section{Experiments}
\label{sec:experiments}
We evaluate the performance of manipulation on Meta-World MT50, RoboTwin, and three real-world tasks on an AgileX dual-arm robot. Meta-World and RoboTwin are reported together in a single simulation table (Table~\ref{tab:1}): Meta-World provides the main quantitative setting across four difficulty tiers, while RoboTwin tests the model in a separate training and deployment setting. Real-world experiments assess performance on a physical robot. 

\subsection{Meta-World and Robotwin Experimental setup}

% \textbf{Meta-World MT50.} Meta-World contains 50 manipulation tasks \citep{Meta-World}. We evaluate 10 episodes per task. Following the aggregation used in the comparison literature \citep{Evo1,FabriVLA}, we report success within each of four difficulty tiers and their unweighted average:

% \[
% \mathrm{SR}_{\mathrm{tier}}
% =\frac{1}{4}\sum_{g=1}^{4}
% \frac{1}{|\mathcal T_g|}\sum_{k\in\mathcal T_g}\mathrm{SR}_k,
% \]

% where $\mathcal T_g$ is the task set in tier $g$. This metric gives equal weight to tiers, whose numbers of tasks may differ. It should therefore be distinguished from an average over all 50 tasks.

% We report the highest recorded tier-averaged score of Evo-World. Inference uses 32 flow-matching integration steps. The Appendix lists the corresponding tier scores.

Evo-World is initialized from general-purpose pretrained encoders and trained on four NVIDIA RTX 4090 GPUs, without additional robot-policy pretraining. Detailed configurations are provided in Appendix Table~\ref{tab:A1}.

\begin{figure}[!htb]
\centering
\includegraphics[width=\linewidth]{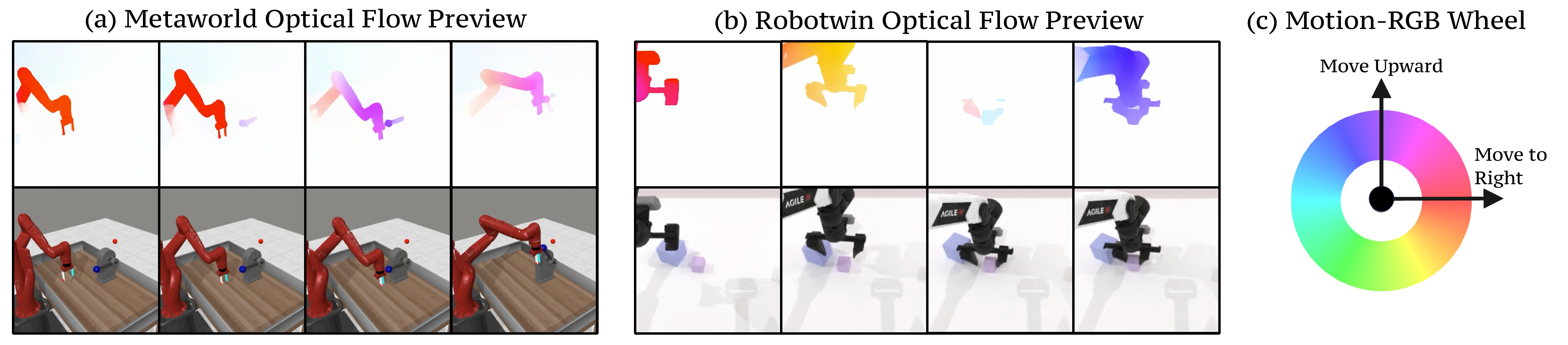}
\caption{\textbf{Optical-flow supervision on Meta-World and RoboTwin.} Panels (a) and (b) pair optical-flow renderings (top) with corresponding RGB observations (bottom) from Meta-World and RoboTwin, respectively. Colored regions emphasize motion, while nearly static regions appear close to white. Panel (c) shows the motion-to-color mapping: hue indicates displacement direction, and weaker motion produces colors closer to white.}
\label{fig:3}
\end{figure}

\textbf{Optical-flow supervision.} Figure~\ref{fig:3} illustrates the optical-flow renderings used to supervise the motion representation module on both simulation benchmarks. Compared with the corresponding RGB observations, these renderings emphasize image-space displacement and suppress static scene appearance. They serve as training targets for latent motion prediction; no optical-flow estimator is required at inference.

\textbf{Baselines.} For Meta-World we compare with $\pi_0$ \citep{Pi0}, SmolVLA \citep{SmolVLA}, RoboTron-Mani \citep{RoboTronMani}, Evo-1 \citep{Evo1}, Evo-Depth \citep{EvoDepth}, LA4VLA \citep{LA4VLA}, and FabriVLA \citep{FabriVLA}. For RoboTwin we compare against DP \citep{DiffusionPolicy}, ACT \citep{ACT}, RDT\citep{rdt}, DP3\citep{dp3}, starVLA \citep{StarVLA}, AHA-WAM\citep{aha}, GalaxeaVLA\citep{GalaxeaVLA}, and Xiaomi Robotics-0 \citep{XiaomiRobotics0}. Table~\ref{tab:1} also lists parameter counts and whether large-scale robot-data pretraining was used. The comparison contextualizes performance.

\subsection{Meta-World and RoboTwin results}

Table~\ref{tab:1} reports both simulation benchmarks. On Meta-World MT50, Evo-World reaches a tier-averaged success rate of 91.1\%, which is 10.5 percentage points above the reported Evo-1 score of 80.6\% and 1.1 points above FabriVLA's 90.0\%. Evo-World obtains the best score on the Hard tier and on the tier average, and the second-best score on the other three tiers, so the four tier scores are consistently high rather than driven by a single difficulty level. On RoboTwin, Evo-World achieves a Clean success rate of 67.1\% and a Random success rate of 16.5\%. Averaged across the Clean and Random settings, Evo-World achieves the highest overall success rate of 41.8\% among the compared methods.
Notably, Evo-World uses about 0.87B parameters and no robot-data pretraining, yet exceeds larger and robot-pretrained baselines such as $\pi_0$ (3.5B) and LA4VLA (1.0B) on the Meta-World average.
Example episode trajectories from the simulation benchmarks are provided in the Appendix~\ref{app:benchmark-examples}.

\begin{table}[!htb]
\setlength{\abovecaptionskip}{0pt} % The caption starts the table float.
\caption{\textbf{Meta-World MT50 and RoboTwin success rates (\%).} Meta-World reports success within four difficulty tiers and their equal-weight tier average; RoboTwin reports the Clean and Randomized settings and their average. Params are in billions, and Pretrain marks large-scale robot-data pretraining. The best score in each column is \textbf{bold} and the second-best is \underline{underlined}; ``\textendash'' denotes a value not reported or not yet evaluated. Evo-World (Ours) is evaluated as described above.}
\vspace{7pt}
\label{tab:1}
\centering\footnotesize
\setlength{\tabcolsep}{3pt}
\renewcommand{\arraystretch}{1.15}
\begin{lrbox}{\papertablebox}
\begin{tabular}{@{}lccrrrrrrrr@{}}
\toprule
& & & \multicolumn{5}{c}{Meta-World MT50} & \multicolumn{3}{c}{RoboTwin} \\
\cmidrule(lr){4-8}\cmidrule(lr){9-11}
Method & Params & Pretrain & Easy & Medium & Hard & Very Hard & Avg & Clean & Random & Avg \\
\midrule
DP & 0.1B & No & 23.1 & 10.7 & 1.9 & 6.1 & 10.5 & 28.0 & 0.6 & 14.3 \\
ACT & 0.08B & No & \textendash & \textendash & \textendash & \textendash & \textendash & 29.7 & 1.7 & 15.7 \\
RDT & 1.2B & Yes & \textendash & \textendash & \textendash & \textendash & \textendash & 34.5 & 13.7 & 24.1 \\
starVLA & 4.68B & No & 85.7 & 60.0 & 46.7 & 44.0 & 59.1 & 46.5 & 3.2 & 24.8 \\
DP3 & 0.26B & No & \textendash & \textendash & \textendash & \textendash & \textendash & 55.2 & 5.0 & 30.1 \\
$\pi_0$ & 3.5B & Yes & 71.8 & 48.2 & 41.7 & 30.0 & 47.9 & 46.4 & 16.3 & 31.4 \\
AHA-WAM & 7.35B & No & \textendash & \textendash & \textendash & \textendash & \textendash & 64.3 & 3.2 & 33.8 \\
GalaxeaVLA & 2.86B & Yes & \textendash & \textendash & \textendash & \textendash & \textendash & 62.7 & 12.7 & 37.7 \\
Xiaomi Robotics-0 & 6.7B & Yes & \textendash & \textendash & \textendash & \textendash & \textendash & 62.9 & \textbf{18.2} & $\underline{40.5}$ \\
SmolVLA & 2.25B & No & 87.1 & 51.8 & 70.0 & 64.0 & 68.2 & \textendash & \textendash & \textendash \\
RoboTron-Mani & 4B & Yes & 85.5 & 67.7 & 76.7 & 81.0 & 77.7 & \textendash & \textendash & \textendash \\
Evo-1 & 0.77B & No & 89.2 & 76.8 & 77.2 & 79.2 & 80.6 & $\underline{65.7}$ & 12.5 & 39.1 \\
Evo-Depth & 0.9B & No & 83.1 & 84.7 & $\underline{87.3}$ & 82.4 & 84.4 & \textendash & \textendash & \textendash \\
LA4VLA & 1.0B & Yes & 88.9 & \textbf{94.5} & 66.7 & \textbf{100.0} & 87.5 & \textendash & \textendash & \textendash \\
FabriVLA & 0.88B & No & \textbf{95.0} & 88.2 & 86.7 & 90.0 & $\underline{90.0}$ & \textendash & \textendash & \textendash \\
\rowcolor{gray!15}
\midrule
\textbf{Evo-World (Ours)} & 0.87B & No & $\underline{93.2}$ & $\underline{89.1}$ & \textbf{88.3} & $\underline{94.0}$ & \textbf{91.1} & \textbf{67.1} & $\underline{16.5}$ & \textbf{41.8} \\
\bottomrule
\end{tabular}
\end{lrbox}
\ifdim\wd\papertablebox>\linewidth
\resizebox{\linewidth}{!}{\usebox{\papertablebox}}
\else
\usebox{\papertablebox}
\fi
\end{table}

\subsection{Real-world experiments}

\begin{figure}[!htb]
\centering
\includegraphics[width=\linewidth]{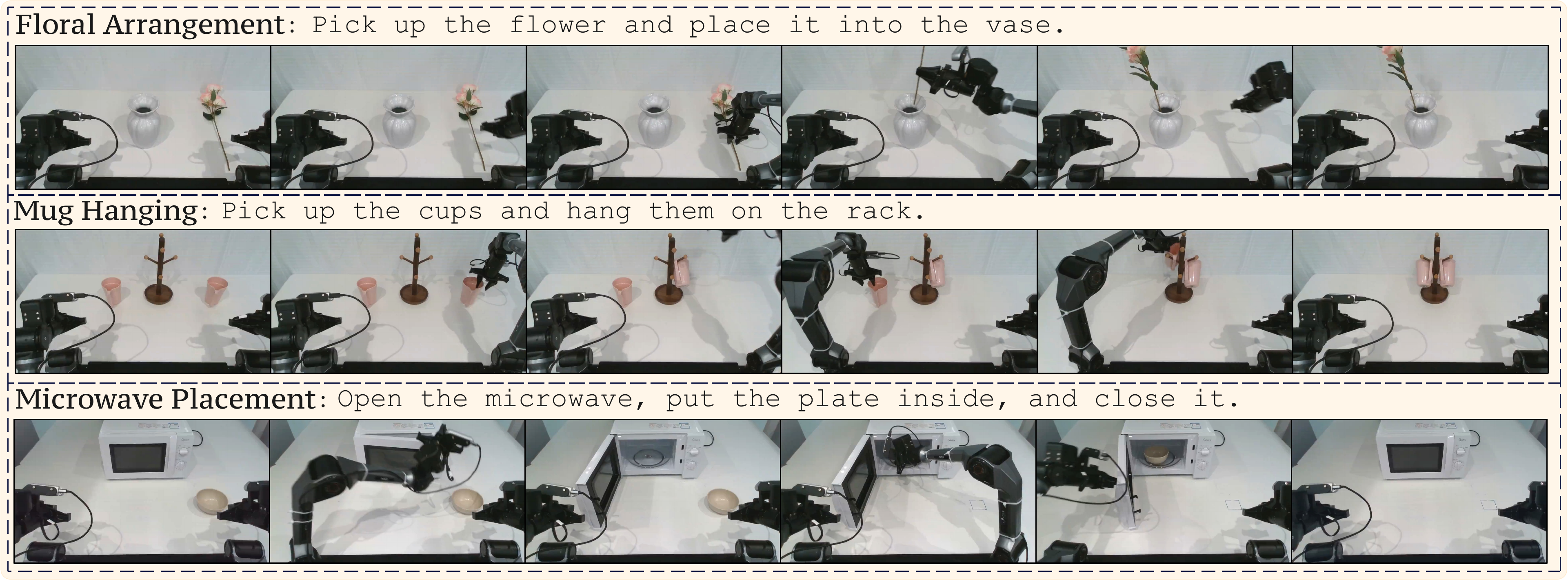}
\caption{\textbf{Successful real-world executions of Evo-World on an AgileX robot.} The rows show floral arrangement (top), a single-arm task, and mug hanging (middle) and microwave placement (bottom), two bimanual collaboration tasks. }
\label{fig:4}
\end{figure}

We evaluate Evo-World on an AgileX dual-arm robot in three tasks covering both single-arm manipulation and bimanual collaboration (Figure~\ref{fig:4}). Floral arrangement is a single-arm task in which one arm inserts a flower stem into a vase. Mug hanging and microwave placement are bimanual collaboration tasks: the two arms work together to hang mugs on a rack or to open a microwave door, put the plate inside, and close the door, respectively. Appendix~\ref{app:real-world-setup} describes the hardware setup, demonstration collection, and supervised fine-tuning protocol.

We compare Evo-World with OpenVLA-OFT \citep{openvlaoft} and $\pi_0$ \citep{Pi0}. Each model is evaluated for 20 trials per task, and we report task success rates and their unweighted mean. All three models run on the same NVIDIA H20 GPU in BF16 precision. We also report parameter count, GPU memory usage, and measured inference frequency for each deployment configuration.

\begin{table}[!htb]
\setlength{\abovecaptionskip}{0pt} % The caption starts the table float.
\caption{\textbf{Real-world performance and deployment efficiency on an AgileX dual-arm robot.} Each model is evaluated for 20 trials per task. All models use the same NVIDIA H20 GPU and BF16 precision. Task success rates and Avg SR are percentages; Avg SR weights the three tasks equally. Lower parameter count and GPU memory usage, and higher inference frequency and success rates, are better. The best value in each row is bold and the second-best is underlined.}
\vspace{7pt}
\label{tab:2}
\centering\footnotesize
\setlength{\tabcolsep}{3pt}
\renewcommand{\arraystretch}{1.15}
\begin{lrbox}{\papertablebox}
\begin{tabular}{@{}lrrr@{}}
\toprule
Metric / Task & OpenVLA-OFT & $\pi_0$ & \textbf{Evo-World (Ours)} \\
\midrule
Params (B) $\downarrow$ & 7.73 & $\underline{3.50}$ & \textbf{0.87} \\
GPU Mem. (GB) $\downarrow$ & 15.56 & $\underline{6.69}$ & \textbf{2.54} \\
Infer. Freq. (Hz) $\uparrow$ & 2.41 & $\underline{5.42}$ & \textbf{6.16} \\
\midrule
Floral arrangement & 25.0 & $\underline{40.0}$ & \textbf{55.0} \\
Mug hanging & 25.0 & $\underline{55.0}$ & \textbf{70.0} \\
Microwave placement & 20.0 & $\underline{60.0}$ & \textbf{65.0} \\
Avg SR & 23.3 & $\underline{51.7}$ & \textbf{63.3} \\
\bottomrule
\end{tabular}
\end{lrbox}
\ifdim\wd\papertablebox>\linewidth
\resizebox{\linewidth}{!}{\usebox{\papertablebox}}
\else
\usebox{\papertablebox}
\fi
\end{table}

Evo-World achieves the highest success rate on all three tasks: 55.0\% for floral arrangement, 70.0\% for mug hanging, and 65.0\% for microwave placement. Its average success rate is 63.3\%, compared with 51.7\% for $\pi_0$ and 23.3\% for OpenVLA-OFT. The gains over $\pi_0$ are 15.0, 15.0, and 5.0 percentage points, respectively, covering both single-arm manipulation and bimanual collaboration.

With 0.87B parameters, Evo-World uses 2.54\,GB of GPU memory and achieves a measured inference frequency of 6.16\,Hz, giving the lowest memory usage and highest inference frequency among the evaluated models. Relative to $\pi_0$, it uses 62.0\% less GPU memory and has a 13.7\% higher inference frequency under the tested deployment settings. These measurements support the practical efficiency of the compact policy alongside its higher task success rates.

% Maintained together with ablation_merged_table.md in this standalone project.

\section{Ablations and Representation Analysis}
\label{sec:ablation}

Table~\ref{tab:3} organizes our Meta-World study along four axes: prediction target (A), FiLM conditioning content (B), fusion mechanism (C), and optimization components (D). Representation contraction and linear decodability of future motion are analyzed in Appendix ~\ref{app:representation-contraction} and Appendix ~\ref{app:motion-decoding}, respectively. All ablation variants are trained and evaluated under the same training budget, using checkpoints after 100,000 second-stage training steps.

% Table 3. Scores and notes are synchronized with ablation_merged_table.md.
% Allow placement here or at the top of a text page; omit the float-page option.
\begin{table}[!ht]
\setlength{\abovecaptionskip}{0pt} % The caption starts the table float.
\caption{\textbf{Ablations on Meta-World MT50 (SR, \%).} Tier Avg. weights the four difficulty buckets equally. Each entry gives absolute SR and, in parentheses, its difference from Evo-World (Ours) in percentage points. Positive, negative, and zero differences are green, red, and gray, respectively. The best and second-best scores in each column are bold and underlined; ties share the same formatting. The shared full model is bold, and differences are computed from the displayed SRs.}
\vspace{7pt}
\label{tab:3}
\centering\footnotesize
\ablSetup
\begin{lrbox}{\papertablebox}
\begin{tabular}{@{}llccccc@{}}
\toprule
ID & Configuration & Easy & Medium & Hard & Very Hard & Tier Avg. \\
\midrule
\multicolumn{7}{@{}l@{}}{\textit{A. Prediction target}} \\
A1 & None & \ablCell{\ablPlain}{89.2}{}{\ablDelta{ablDarkRed}{-}{4.0}} & \ablCell{\ablPlain}{76.8}{}{\ablDelta{ablDarkRed}{-}{12.3}} & \ablCell{\ablPlain}{77.2}{}{\ablDelta{ablDarkRed}{-}{11.1}} & \ablCell{\ablPlain}{79.2}{}{\ablDelta{ablDarkRed}{-}{14.8}} & \ablCell{\ablPlain}{80.6}{}{\ablDelta{ablDarkRed}{-}{10.5}} \\
A2 & Future RGB & \ablCell{\underline}{90.7}{}{\ablDelta{ablDarkRed}{-}{2.5}} & \ablCell{\underline}{80.0}{}{\ablDelta{ablDarkRed}{-}{9.1}} & \ablCell{\underline}{78.3}{}{\ablDelta{ablDarkRed}{-}{10.0}} & \ablCell{\underline}{88.0}{}{\ablDelta{ablDarkRed}{-}{6.0}} & \ablCell{\underline}{84.3}{}{\ablDelta{ablDarkRed}{-}{6.8}} \\
\rowcolor{gray!15}
A3 & \textbf{Optical flow (Ours)} & \ablCell{\textbf}{93.2}{}{\ablSelf} & \ablCell{\textbf}{89.1}{}{\ablSelf} & \ablCell{\textbf}{88.3}{}{\ablSelf} & \ablCell{\textbf}{94.0}{}{\ablSelf} & \ablCell{\textbf}{91.1}{}{\ablSelf} \\
\midrule
\multicolumn{7}{@{}l@{}}{\textit{B. FiLM conditioning content}} \\
B1 & $z_t$ & \ablCell{\ablPlain}{89.2}{}{\ablDelta{ablDarkRed}{-}{4.0}} & \ablCell{\underline}{80.9}{}{\ablDelta{ablDarkRed}{-}{8.2}} & \ablCell{\ablPlain}{76.7}{}{\ablDelta{ablDarkRed}{-}{11.6}} & \ablCell{\underline}{88.0}{}{\ablDelta{ablDarkRed}{-}{6.0}} & \ablCell{\ablPlain}{83.7}{}{\ablDelta{ablDarkRed}{-}{7.4}} \\
B2 & $z_{\mathrm{pred}}$ & \ablCell{\underline}{90.4}{}{\ablDelta{ablDarkRed}{-}{2.8}} & \ablCell{\ablPlain}{79.0}{}{\ablDelta{ablDarkRed}{-}{10.1}} & \ablCell{\underline}{86.7}{}{\ablDelta{ablDarkRed}{-}{1.6}} & \ablCell{\ablPlain}{86.0}{}{\ablDelta{ablDarkRed}{-}{8.0}} & \ablCell{\underline}{85.5}{}{\ablDelta{ablDarkRed}{-}{5.6}} \\
B3 & Noise & \ablCell{\ablPlain}{86.5}{}{\ablDelta{ablDarkRed}{-}{6.7}} & \ablCell{\ablPlain}{77.3}{}{\ablDelta{ablDarkRed}{-}{11.8}} & \ablCell{\ablPlain}{76.0}{}{\ablDelta{ablDarkRed}{-}{12.3}} & \ablCell{\ablPlain}{84.0}{}{\ablDelta{ablDarkRed}{-}{10.0}} & \ablCell{\ablPlain}{81.0}{}{\ablDelta{ablDarkRed}{-}{10.1}} \\
\rowcolor{gray!15}
B4 & $\boldsymbol{z_t} + \boldsymbol{z_{\mathrm{pred}}}$ \textbf{(Ours)} & \ablCell{\textbf}{93.2}{}{\ablSelf} & \ablCell{\textbf}{89.1}{}{\ablSelf} & \ablCell{\textbf}{88.3}{}{\ablSelf} & \ablCell{\textbf}{94.0}{}{\ablSelf} & \ablCell{\textbf}{91.1}{}{\ablSelf} \\
\midrule
\multicolumn{7}{@{}l@{}}{\textit{C. Fusion mechanism}} \\
C1 & Concatenation & \ablCell{\ablPlain}{81.1}{}{\ablDelta{ablDarkRed}{-}{12.1}} & \ablCell{\underline}{37.3}{}{\ablDelta{ablDarkRed}{-}{51.8}} & \ablCell{\ablPlain}{48.3}{}{\ablDelta{ablDarkRed}{-}{40.0}} & \ablCell{\ablPlain}{46.0}{}{\ablDelta{ablDarkRed}{-}{48.0}} & \ablCell{\ablPlain}{53.2}{}{\ablDelta{ablDarkRed}{-}{37.9}} \\
C2 & Cross-attention & \ablCell{\underline}{92.1}{}{\ablDelta{ablDarkRed}{-}{1.1}} & \ablCell{\textbf}{89.1}{}{\ablDelta{gray}{}{0.0}} & \ablCell{\underline}{78.3}{}{\ablDelta{ablDarkRed}{-}{10.0}} & \ablCell{\underline}{86.0}{}{\ablDelta{ablDarkRed}{-}{8.0}} & \ablCell{\underline}{86.4}{}{\ablDelta{ablDarkRed}{-}{4.7}} \\
\rowcolor{gray!15}
C3 & \textbf{FiLM (Ours)} & \ablCell{\textbf}{93.2}{}{\ablSelf} & \ablCell{\textbf}{89.1}{}{\ablSelf} & \ablCell{\textbf}{88.3}{}{\ablSelf} & \ablCell{\textbf}{94.0}{}{\ablSelf} & \ablCell{\textbf}{91.1}{}{\ablSelf} \\
\midrule
\multicolumn{7}{@{}l@{}}{\textit{D. Component ablation}} \\
D1 & w/o SIGReg & \ablCell{\ablPlain}{91.0}{}{\ablDelta{ablDarkRed}{-}{2.2}} & \ablCell{\ablPlain}{87.3}{}{\ablDelta{ablDarkRed}{-}{1.8}} & \ablCell{\ablPlain}{86.7}{}{\ablDelta{ablDarkRed}{-}{1.6}} & \ablCell{\ablPlain}{78.0}{}{\ablDelta{ablDarkRed}{-}{16.0}} & \ablCell{\ablPlain}{85.8}{}{\ablDelta{ablDarkRed}{-}{5.3}} \\
D2 & w/o Grad. Detach & \ablCell{\ablPlain}{92.1}{}{\ablDelta{ablDarkRed}{-}{1.1}} & \ablCell{\ablPlain}{84.6}{}{\ablDelta{ablDarkRed}{-}{4.5}} & \ablCell{\ablPlain}{68.3}{}{\ablDelta{ablDarkRed}{-}{20.0}} & \ablCell{\ablPlain}{84.0}{}{\ablDelta{ablDarkRed}{-}{10.0}} & \ablCell{\ablPlain}{82.2}{}{\ablDelta{ablDarkRed}{-}{8.9}} \\
\rowcolor{gray!15}
D3 & \textbf{Ours} & \ablCell{\textbf}{93.2}{}{\ablSelf} & \ablCell{\textbf}{89.1}{}{\ablSelf} & \ablCell{\textbf}{88.3}{}{\ablSelf} & \ablCell{\textbf}{94.0}{}{\ablSelf} & \ablCell{\textbf}{91.1}{}{\ablSelf} \\
\bottomrule
\end{tabular}
\end{lrbox}
\ablUseTable
\par\smallskip
\begin{minipage}{\linewidth}
\footnotesize\raggedright
\textit{Shared model:} A3, B4, C3, and D3 repeat the same Evo-World (Ours) result, using optical-flow supervision, current and predicted representations, and FiLM; \textemdash{} marks a self-comparison.
\end{minipage}
\end{table}

\subsection{Comparing prediction targets}

Group A compares None prediction baseline, the future-frame-supervised variant, and the optical-flow-supervised model. The future-frame variant predicts an encoding of a future RGB image; the optical-flow variant predicts an encoding of rendered displacement. Neither variant reconstructs image pixels. 

The future-frame variant exceeds the baseline result by 3.7 percentage points, and Evo-World further improves upon the future-frame variant by 6.8 points. This ordering supports optical-flow supervision among the evaluated configurations. Since all variants use the same training and evaluation protocol, these differences isolate the effect of the prediction target.

\subsection{FiLM conditioning content under optical-flow supervision}

Group B evaluates which representations should condition FiLM under optical-flow supervision. We compare the current representation $z_t$ alone, the predicted representation $z_{\mathrm{pred}}$ alone, a noise condition, and the joint condition $z_t + z_{\mathrm{pred}}$. Here, the joint condition denotes concatenation, and B4 repeats the full-model result.

Joint conditioning achieves 91.1\% tier-averaged success, exceeding the current-only (83.7\%), prediction-only (85.5\%), and noise (81.0\%) conditions by 7.4, 5.6, and 10.1 percentage points, respectively. It also achieves the highest success rate in all four difficulty buckets. Prediction-only conditioning exceeds current-only conditioning by 1.8 points on average, with higher scores on Easy and Hard; current-only conditioning performs better on Medium and Very Hard. The joint condition's consistent gains over both individual representations support combining current scene context with anticipated motion to condition action generation. The corresponding future-frame-supervised comparisons are reported separately in Appendix~\ref{app:future-frame-fusion}.

\subsection{Comparing fusion mechanisms}

Group C compares concatenation, cross-attention, and FiLM for fusing motion representations with the policy's vision-language features. The FiLM row reproduces the highest recorded full-model result reported in Table~\ref{tab:1}.

FiLM achieves 91.1\% tier-averaged success, exceeding concatenation and cross-attention by 37.9 and 4.7 percentage points, respectively. It obtains the highest reported score in the Easy, Hard, and Very Hard buckets and ties cross-attention in Medium. These results support the choice of FiLM among the reported fusion configurations. Earlier fusion comparisons under future-frame supervision are retained in Appendix ~\ref{app:future-frame-fusion}.

\subsection{Effects of regularization and gradient detachment}

Group D compares the full model with variants that remove SIGReg or gradient detachment. The component study uses the same two-stage training schedule, optical-flow cache, and evaluation settings. Each variant changes the corresponding component throughout training. Absolute SRs are obtained by adding the reported component-ablation differences to the corresponding full-model scores.

Removing SIGReg reduces the overall success rate by 5.3 percentage points, with performance declining across all difficulty tiers. Thus, regularization improves aggregate performance consistently across all four difficulty tiers. Further analysis of the effect of the SIGReg loss on the learned representations is provided in the Appendix ~\ref{app:representation-contraction}.

Removing gradient detachment reduces success by 8.9 points, including a 20.0-point decline in the Hard tier. This result supports retaining separate gradient paths for motion representation learning and action optimization in the tested configuration.

% Generated from content/7_Conclusion.md by review/sync_tex.py. Edit the Markdown source.

\section{Conclusion}

We presented \textbf{Evo-World}, a compact VLA policy that learns instruction-conditioned predictions of future motion through optical-flow supervision. By directly predicting learned embeddings of flow renderings and using them to condition action generation, the model introduces motion anticipation without reconstructing future images or optical flow. 
% Distribution regularization and gradient separation support the motion representations during policy learning, while current and predicted features jointly guide the action expert.

Experiments on Meta-World MT50, RoboTwin, and an AgileX robot support the effectiveness of this formulation. With \textbf{0.87B} parameters, it outperforms the evaluated real-world baselines across single-arm and bimanual tasks while using less GPU memory and achieving a higher measured inference frequency. Ablations further support joint conditioning on current and predicted representations, distribution regularization, and gradient separation. Because the current motion targets describe 2D image-plane displacement without explicit 3D geometry, future work will explore geometry-aware supervision while preserving efficient deployment.

\FloatBarrier

\bibliography{iclr2027_conference}
\bibliographystyle{iclr2027_conference}

\clearpage
\appendix
\makeatletter
\setlength{\@fptop}{0pt}
\makeatother
\renewcommand{\theHtable}{appendix.\arabic{table}}
\renewcommand{\theHfigure}{appendix.\arabic{figure}}
\setcounter{table}{0}
\renewcommand{\thetable}{A\arabic{table}}
\setcounter{figure}{0}
\renewcommand{\thefigure}{A\arabic{figure}}
% Maintained directly in this standalone project; no local Markdown source.

\section{Implementation details}
\label{app:implementation}
\subsection{Representative training hyperparameters on Meta-World}

Table~\ref{tab:A1} summarizes the two-stage Meta-World training configuration. Stage 1 trains the motion representation module, FiLM, and action expert with the vision-language backbone frozen. Stage 2 initializes from the Stage 1 model, unfreezes the backbone, and starts a new optimizer and learning-rate schedule for a further 100,000 steps. Both motion encoders, their projectors, and the motion predictor remain trainable in both stages. The global batch contains 64 samples: 16 per GPU on four NVIDIA RTX 4090 GPUs, with one optimizer update per batch.

% Representative Meta-World settings; stage values follow the training launch configuration.
\begin{table}[!ht]
\caption{\textbf{Representative Training Hyperparameters on Meta-World.} Evo-World (Ours) uses two training stages. Entries spanning both stage columns apply to both stages; the loss coefficients follow Eq.~(\ref{eq:method-5}).}
\vspace{7pt}
\label{tab:A1}
\centering\footnotesize
\setlength{\tabcolsep}{5pt}
\renewcommand{\arraystretch}{1.12}
\begin{lrbox}{\papertablebox}
\begin{tabular}{@{}lcc@{}}
\toprule
\textbf{Hyperparameter} & \textbf{Stage 1} & \textbf{Stage 2} \\
\midrule
\multicolumn{3}{@{}l@{}}{\textbf{\textit{Training stages}}} \\
Training steps & 10,000 & 100,000 \\
Vision-language backbone & Frozen & Trainable \\
Motion representation module & Trainable & Trainable \\
FiLM and action expert & Trainable & Trainable \\
\midrule
\multicolumn{3}{@{}l@{}}{\textbf{\textit{Optimization}}} \\
Optimizer & \multicolumn{2}{c@{}}{AdamW} \\
Peak learning rate & \multicolumn{2}{c@{}}{$10^{-5}$} \\
Learning-rate schedule & \multicolumn{2}{c@{}}{Linear warmup + cosine decay} \\
Warmup steps per stage & \multicolumn{2}{c@{}}{1,000} \\
Weight decay & \multicolumn{2}{c@{}}{$10^{-3}$} \\
Adam coefficients $(\beta_1,\beta_2)$ & \multicolumn{2}{c@{}}{$(0.9,0.999)$} \\
Adam $\epsilon$ & \multicolumn{2}{c@{}}{$10^{-8}$} \\
Gradient clipping (maximum norm) & \multicolumn{2}{c@{}}{1.0} \\
Batch size per GPU & \multicolumn{2}{c@{}}{16} \\
Effective global batch size & \multicolumn{2}{c@{}}{64} \\
Gradient accumulation steps & \multicolumn{2}{c@{}}{1} \\
\midrule
\multicolumn{3}{@{}l@{}}{\textbf{\textit{Motion conditioning and regularization}}} \\
World-loss weight $\lambda_{\mathrm{world}}$ & \multicolumn{2}{c@{}}{0.1} \\
SIGReg coefficient $\lambda_{\mathrm{sig}}$ & \multicolumn{2}{c@{}}{0.09} \\
FiLM scale $\alpha$ & \multicolumn{2}{c@{}}{0.1} \\
Detachment at the policy interface & \multicolumn{2}{c@{}}{Enabled} \\
Action expert dropout & \multicolumn{2}{c@{}}{0.2} \\
\midrule
\multicolumn{3}{@{}l@{}}{\textbf{\textit{Runtime and input settings}}} \\
Training GPUs & \multicolumn{2}{c@{}}{$4\times$ NVIDIA RTX 4090} \\
Mixed precision & \multicolumn{2}{c@{}}{BF16} \\
Distributed training & \multicolumn{2}{c@{}}{DeepSpeed ZeRO stage 2} \\
Activation checkpointing & \multicolumn{2}{c@{}}{Enabled} \\
Policy image resolution & \multicolumn{2}{c@{}}{$448\times448$} \\
Motion-branch image resolution & \multicolumn{2}{c@{}}{$224\times224$} \\
Action chunk length & \multicolumn{2}{c@{}}{50} \\
Policy augmentation probability & \multicolumn{2}{c@{}}{0.5 per view} \\
\bottomrule
\end{tabular}
\end{lrbox}
\ifdim\wd\papertablebox>\linewidth
\resizebox{\linewidth}{!}{\usebox{\papertablebox}}
\else
\usebox{\papertablebox}
\fi
\end{table}

Each stage uses 1,000 steps of linear learning-rate warmup followed by cosine decay. Biases, normalization parameters, and other one-dimensional parameters are excluded from weight decay. Policy-image augmentation is applied independently with probability 0.5 per view and combines a random resized crop with area scale $[0.95,1.0]$, rotation within $[-5^{\circ},5^{\circ}]$, and color jitter with brightness, contrast, saturation, and hue strengths of 0.3, 0.4, 0.5, and 0.08, respectively. Current images and optical-flow targets in the motion branch use the deterministic preprocessing described below.

\subsection{Exact SIGReg statistic}
\label{app:sigreg_sta}
We instantiate the characteristic-function form of SIGReg \citep{LeJEPA} as follows. For each local training batch, sample $P=1024$ independent unit directions $u_p\in\mathbb R^D$ by normalizing Gaussian vectors. Use $K=17$ knots $\xi_k=3(k-1)/(K-1)$ and $\delta=3/(K-1)$. Define $q_1=q_K=\delta$ and $q_k=2\delta$ for interior knots. The implementation computes

\begin{multline*}
\mathcal S(Z)
=\frac{B}{P}\sum_{p=1}^{P}\sum_{k=1}^{K}
q_k e^{-\xi_k^2/2}
\\\times\left[
\left(\frac{1}{B}\sum_{b=1}^{B}\cos(\xi_k u_p^\top z_b)-e^{-\xi_k^2/2}\right)^2
+\left(\frac{1}{B}\sum_{b=1}^{B}\sin(\xi_k u_p^\top z_b)\right)^2
\right].
\end{multline*}

These $q_k$ are twice the conventional trapezoidal weights on $[0,3]$, equivalently accounting for the symmetric integral on $[-3,3]$. The code averages $\mathcal S(z_t)$ and $\mathcal S(z_h)$ and uses the same sampled directions for the two representations. Characteristic-function averages are computed within the local batch; the world-loss implementation does not gather embeddings across workers.

\subsection{Optical-flow targets}

The Meta-World preprocessing script computes RAFT flow from each source frame to the frame 15 steps ahead. Near an episode boundary, the destination is clipped to the final frame. The default renderer uses a magnitude scale fixed within each dataset, clips larger magnitudes while retaining direction, and saves the renderings in a video with the original frame indexing. Scale calibration, when not supplied explicitly, uses the 99th percentile of sampled magnitudes from the first five episodes. The exact cached scale and RAFT checkpoint should be obtained from the preprocessing manifest.

At training time, the dataset loads the cached flow rendering at the \emph{current} frame timestamp. A new prediction interval requires a corresponding cache.

\subsection{RoboTwin Evaluation Protocol}

The combined comparison in Table~\ref{tab:1} includes Evo-1 \citep{Evo1}, starVLA \citep{StarVLA}, $\pi_0$ \citep{Pi0}, ACT \citep{ACT}, and Xiaomi Robotics-0 \citep{XiaomiRobotics0}, among other baselines. We report results under the aligned RoboTwin Clean2Clean and Clean2Random settings, following the RoboTwin 2.0 evaluation protocol . 
\section{Earlier future-frame fusion experiments}
\label{app:future-frame-fusion}

The earlier future-frame-supervised study reports the results in Table~\ref{tab:A2}. These use appearance-target representations; Table~\ref{tab:3}, group C, reports the separate fusion comparison under optical-flow supervision. Within this earlier study, FiLM and the combined current/predicted condition obtain the highest recorded values. 

% Earlier future-frame results, updated with the author's corrected bucket scores.
\begin{table}[!ht]
\caption{\textbf{Fusion and conditioning comparisons for future-frame supervision (SR, \%).} Groups A and B change different factors. Each entry gives absolute SR and its difference from the corresponding reference score in percentage points. Positive, negative, and zero differences are dark green, dark red, and gray, respectively. Within each group and column, the best score is bold and the second-best is underlined; ties share the same formatting.}
\vspace{7pt}
\label{tab:A2}
\centering\footnotesize
\ablSetup
\begin{lrbox}{\papertablebox}
\begin{tabular}{@{}llccccc@{}}
\toprule
ID & Configuration & Easy & Medium & Hard & Very Hard & Tier Avg. \\
\midrule
\multicolumn{7}{@{}l@{}}{\textit{A. Fusion mechanism}} \\
A1 & Concatenation & \ablCell{\ablPlain}{88.6}{}{\ablDelta{ablDarkRed}{-}{2.1}} & \ablCell{\underline}{79.1}{}{\ablDelta{ablDarkRed}{-}{0.9}} & \ablCell{\ablPlain}{71.7}{}{\ablDelta{ablDarkRed}{-}{6.6}} & \ablCell{\underline}{86.0}{}{\ablDelta{ablDarkRed}{-}{2.0}} & \ablCell{\ablPlain}{81.4}{}{\ablDelta{ablDarkRed}{-}{2.9}} \\
A2 & Cross-attention & \ablCell{\underline}{89.6}{}{\ablDelta{ablDarkRed}{-}{1.1}} & \ablCell{\ablPlain}{74.5}{}{\ablDelta{ablDarkRed}{-}{5.5}} & \ablCell{\textbf}{80.0}{}{\ablDelta{ablDarkGreen}{+}{1.7}} & \ablCell{\underline}{86.0}{}{\ablDelta{ablDarkRed}{-}{2.0}} & \ablCell{\underline}{82.5}{}{\ablDelta{ablDarkRed}{-}{1.7}} \\
\rowcolor{gray!15}
A3 & \textbf{FiLM} & \ablCell{\textbf}{90.7}{}{\ablSelf} & \ablCell{\textbf}{80.0}{}{\ablSelf} & \ablCell{\underline}{78.3}{}{\ablSelf} & \ablCell{\textbf}{88.0}{}{\ablSelf} & \ablCell{\textbf}{84.3}{}{\ablSelf} \\
\midrule
\multicolumn{7}{@{}l@{}}{\textit{B. FiLM conditioning content}} \\
B1 & $z_t$ only & \ablCell{\underline}{89.3}{}{\ablDelta{ablDarkRed}{-}{1.4}} & \ablCell{\underline}{78.2}{}{\ablDelta{ablDarkRed}{-}{1.8}} & \ablCell{\underline}{73.3}{}{\ablDelta{ablDarkRed}{-}{5.0}} & \ablCell{\underline}{86.0}{}{\ablDelta{ablDarkRed}{-}{2.0}} & \ablCell{\underline}{81.7}{}{\ablDelta{ablDarkRed}{-}{2.6}} \\
B2 & Noise & \ablCell{\underline}{89.3}{}{\ablDelta{ablDarkRed}{-}{1.4}} & \ablCell{\ablPlain}{76.4}{}{\ablDelta{ablDarkRed}{-}{3.6}} & \ablCell{\ablPlain}{71.7}{}{\ablDelta{ablDarkRed}{-}{6.6}} & \ablCell{\ablPlain}{78.0}{}{\ablDelta{ablDarkRed}{-}{10.0}} & \ablCell{\ablPlain}{78.9}{}{\ablDelta{ablDarkRed}{-}{5.4}} \\
\rowcolor{gray!15}
B3 & $\boldsymbol{z_t} + \boldsymbol{z_{\mathrm{pred}}}$ & \ablCell{\textbf}{90.7}{}{\ablSelf} & \ablCell{\textbf}{80.0}{}{\ablSelf} & \ablCell{\textbf}{78.3}{}{\ablSelf} & \ablCell{\textbf}{88.0}{}{\ablSelf} & \ablCell{\textbf}{84.3}{}{\ablSelf} \\
\bottomrule
\end{tabular}
\end{lrbox}
\ablUseTable
\par\smallskip
\begin{minipage}{\linewidth}
\footnotesize\raggedright
\textit{Reference:} A3 and B3 repeat the same future-frame baseline, using FiLM with current and predicted representations; \textemdash{} denotes a self-comparison. The joint condition $z_t + z_{\mathrm{pred}}$ denotes concatenation.
\end{minipage}
\end{table}

% \section{RoboTwin evaluation protocol}
% \label{app:robotwin-evaluation}

% The combined comparison in Table~\ref{tab:1} includes Evo-1 \citep{Evo1}, starVLA \citep{StarVLA}, $\pi_0$ \citep{Pi0}, ACT \citep{ACT}, and Xiaomi Robotics-0 \citep{XiaomiRobotics0}, among other baselines. We report results under the aligned RoboTwin Clean2Clean and Clean2Random settings, following the RoboTwin 2.0 evaluation protocol \citep{RoboTwinLeaderboard}. All reported results use the same evaluation settings, enabling direct comparison across methods.

\section{Representation contraction without SIGReg}
\label{app:representation-contraction}

\begin{figure}[htbp]
\centering
\includegraphics[width=\linewidth]{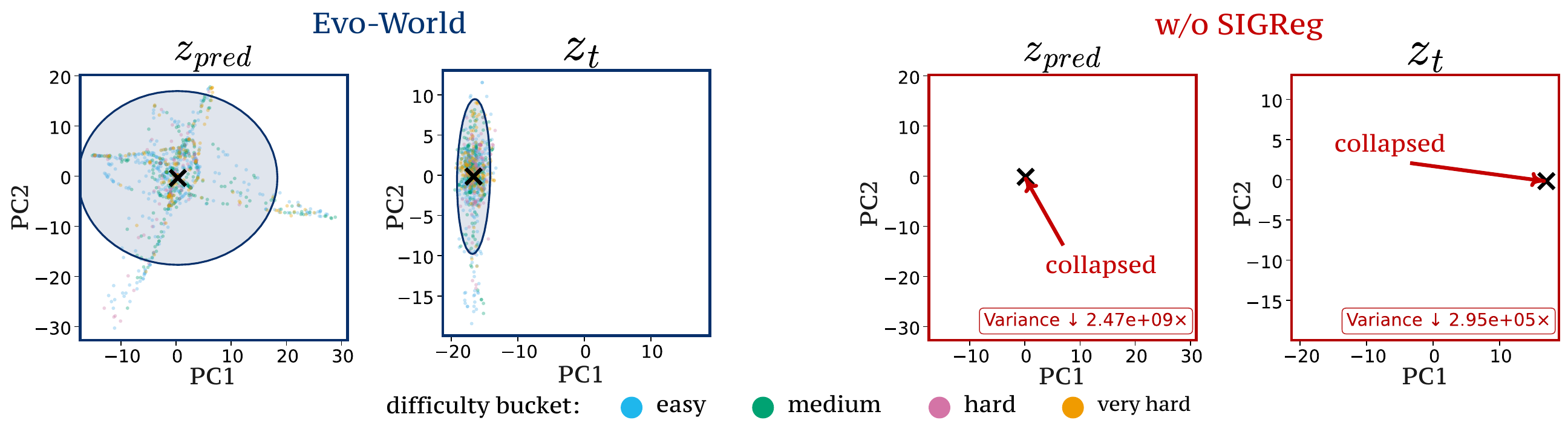}
\caption{\textbf{Effect of SIGReg on latent representations.} PCA projections for Evo-World (left pair) and the variant without SIGReg (right pair). Each pair shows $z_{\mathrm{pred}}$ followed by $z_t$. Colors indicate task difficulty, and black crosses mark the plotted means. For each representation, the corresponding panels use the same displayed axis scales. The annotations report the variance reduction factors when SIGReg is removed.}
\label{fig:A1}
\end{figure}

To examine how SIGReg affects the learned motion representations, Figure~\ref{fig:A1} compares PCA projections of the current representation $z_t$ and predicted representation $z_{\mathrm{pred}}$ with and without regularization. With SIGReg, both representations exhibit variation across observations. Without SIGReg, they concentrate near their respective means, indicating collapse toward nearly constant representations.

PC1 and PC2 denote coordinates along the first and second principal components, respectively; their values are latent-space coordinates, not physical distances. For each representation, the panels with and without SIGReg have the same displayed axis ranges and tick spacing. The $z_{\mathrm{pred}}$ panels use 10-unit tick intervals on both axes, whereas the $z_t$ panels use 10-unit intervals on PC1 and 5-unit intervals on PC2. Representation spread should therefore be compared between regularization conditions within the corresponding pair of panels.

As annotated in Figure~\ref{fig:A1}, removing SIGReg reduces the variance of $z_t$ by a factor of $2.95\times10^5$ and that of $z_{\mathrm{pred}}$ by a factor of $2.47\times10^9$. The contraction affects both representations and is substantially stronger in the predicted representation.

These observations complement the component ablation in Table~\ref{tab:3}, group D, where removing SIGReg lowers tier-averaged success from 91.1\% to 85.8\%, a decrease of 5.3 percentage points. Together, the representation analysis and control results support the role of SIGReg in maintaining input-dependent motion representations for action generation.

\section{Linear decodability of future motion}
\label{app:motion-decoding}

% AUTHOR CHECK P04: The values and evaluation protocol were transcribed from the supplied fig4 optical vs futrue.pdf. Raw features, sample IDs, the linear-decoding implementation, and split-level results were not available in the reviewed directories. Confirm these before treating the subsection as finalized experimental evidence.

Table~\ref{tab:A4} compares how well current and predicted representations encode future end-effector displacement. The reported evaluation keeps the representation models frozen and fits a linear ridge-regression decoder to displacement over 15 frames, using common samples and trajectory-level splits. Displacement labels are derived from simulator state, independently of the optical-flow estimator used for training. Performance is measured by test $R^2$ and median angular error.

\begin{table}[htbp]
\caption{\textbf{Linear decoding of future end-effector displacement.} Results compare current and predicted representations under future-frame and optical-flow supervision. Higher test $R^2$ and lower median angular error indicate better decoding. The best score in each metric column is bold and the second-best is underlined.}
\vspace{7pt}
\label{tab:A4}
\centering\footnotesize
\setlength{\tabcolsep}{3pt}
\renewcommand{\arraystretch}{1.15}
\begin{lrbox}{\papertablebox}
\begin{tabular}{@{}llrr@{}}
\toprule
Supervision & Representation & Test $R^2$ & Median error (deg.) \\
\midrule
Future frame & $z_t$ & 0.497 & 29.6 \\
Future frame & $z_{\mathrm{pred}}$ & 0.421 & 35.2 \\
Optical flow & $z_t$ & \textbf{0.661} & \textbf{20.9} \\
Optical flow & $z_{\mathrm{pred}}$ & $\underline{0.508}$ & $\underline{29.4}$ \\
\bottomrule
\end{tabular}
\end{lrbox}
\ifdim\wd\papertablebox>\linewidth
\resizebox{\linewidth}{!}{\usebox{\papertablebox}}
\else
\usebox{\papertablebox}
\fi
\end{table}

Optical-flow supervision yields higher test $R^2$ and lower median angular error for both current and predicted representations than future-frame supervision. These results indicate that future end-effector displacement is more accessible to linear decoding in the optical-flow-supervised representations. Under both supervision targets, $z_t$ yields better decoding results than $z_{\mathrm{pred}}$. This comparison assesses end-effector displacement and does not establish that the predictor adds linearly decodable motion information beyond the current representation.

\section{Simulation benchmark examples}
\label{app:benchmark-examples}

\begin{figure}[htbp]
\centering
\includegraphics[width=0.8\linewidth]{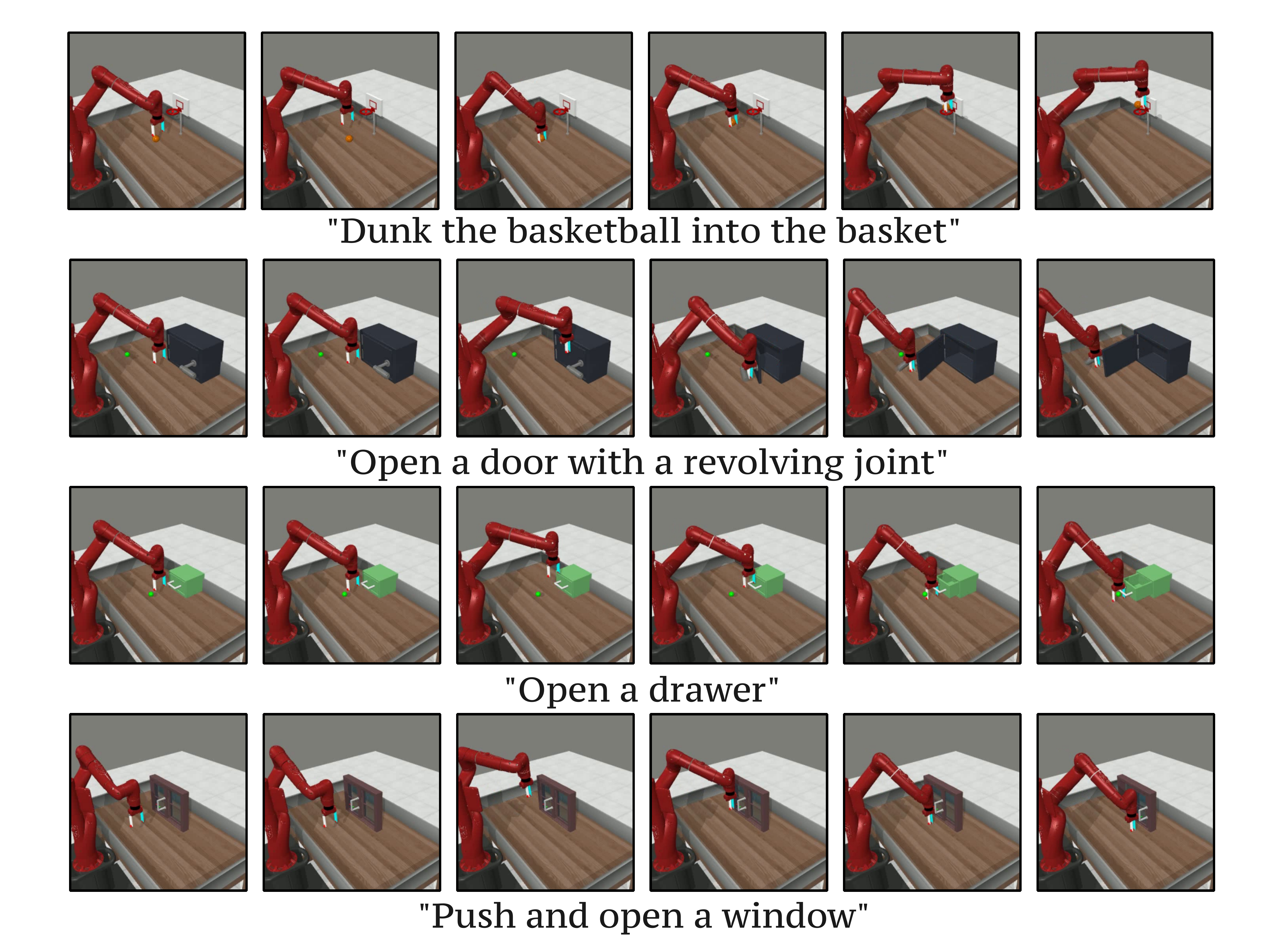}
\caption{\textbf{Example task sequences from Meta-World.} From top to bottom: placing a basketball into a basket, opening a hinged door, opening a drawer, and opening a window. Each row shows six successive observations from left to right, with the corresponding language instruction below.}
\label{fig:Meta-World-examples}
\end{figure}

\begin{figure}[htbp]
\centering
\includegraphics[width=\linewidth]{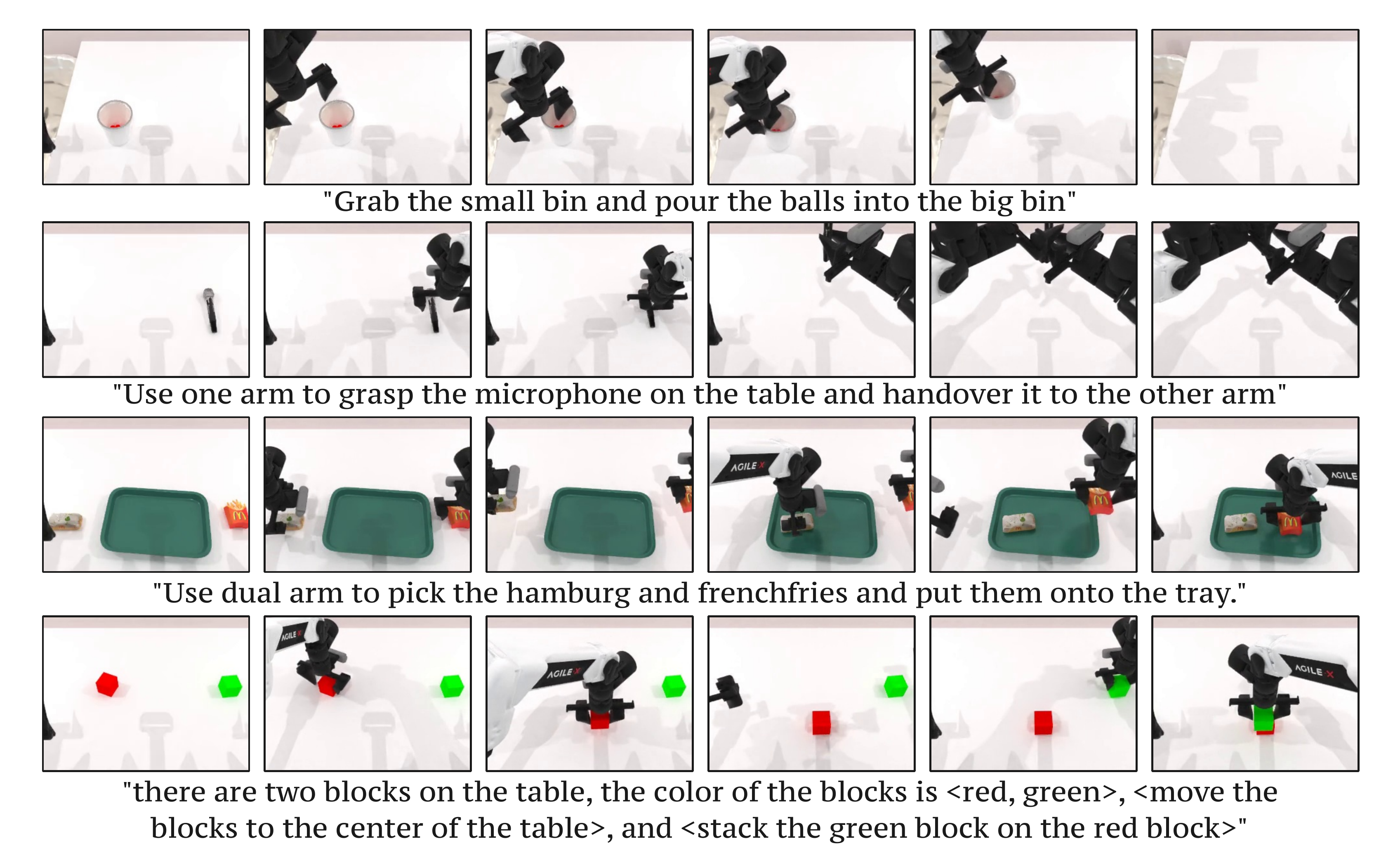}
\caption{\textbf{Example task sequences from RoboTwin.} From top to bottom: pouring balls from a small container into a larger one, handing a microphone from one arm to the other, placing food items on a tray with both arms, and stacking a green block on a red block. Each row shows six successive observations from left to right, with the corresponding language instruction below.}
\label{fig:robotwin-examples}
\end{figure}

Figures~\ref{fig:Meta-World-examples} and~\ref{fig:robotwin-examples} provide example task sequences from Meta-World and RoboTwin, complementing the quantitative comparison in Table~\ref{tab:1}. The Meta-World examples illustrate object transport and interactions with articulated objects, including doors, drawers, and windows. The RoboTwin examples cover container manipulation, object handover between arms, coordinated placement, and stacking. Together, they illustrate the range of object motions and arm coordination involved in the evaluated benchmarks.

\section{Real-world experimental setup}
\label{app:real-world-setup}

\begin{figure}[!t]
\centering
\includegraphics[width=\linewidth]{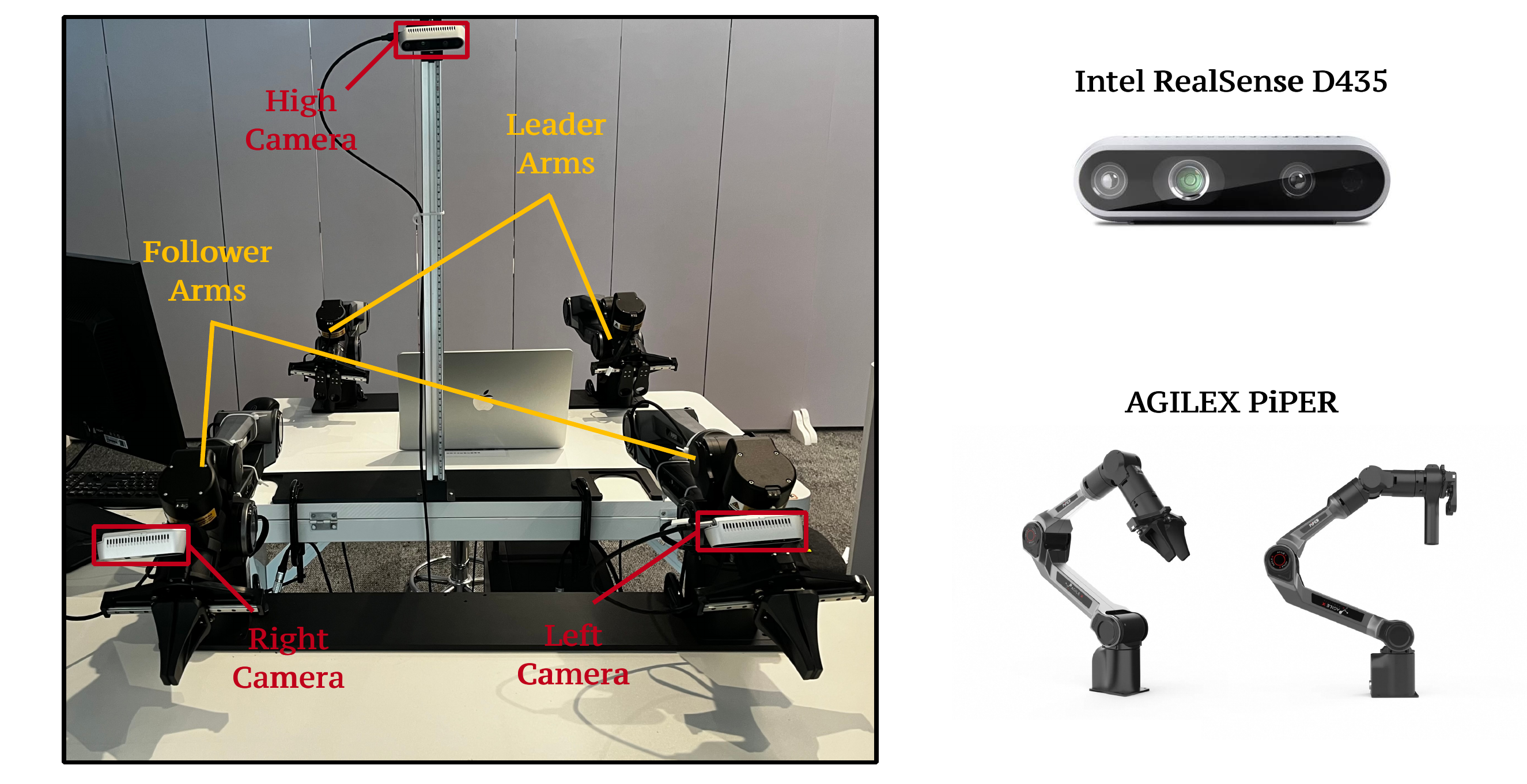}
\caption{\textbf{Real-world experimental setup.} Left: the teleoperation apparatus, with leader arms controlling the two follower arms and cameras providing an elevated view and left and right wrist views. Right: the Intel RealSense D435 camera and AgileX PiPER robot used in the setup. We collect 100 teleoperated demonstrations per task with varied object positions and use the collected data to separately fine-tune OpenVLA-OFT, $\pi_0$, and Evo-World.}
\label{fig:real-world-setup}
\end{figure}

Figure~\ref{fig:real-world-setup} shows the hardware used for real-world data collection and policy evaluation. The setup comprises two AgileX PiPER follower arms, a pair of leader arms for teleoperation, and Intel RealSense D435 cameras providing one elevated view and two wrist-mounted views of the workspace.

For each of the three tasks---floral arrangement, mug hanging, and microwave placement---we collect 100 teleoperated demonstration trajectories, yielding 300 demonstrations in total. Task-relevant object positions are varied across demonstrations to support generalization across spatial configurations. Floral arrangement is performed with a single arm, while mug hanging and microwave placement require bimanual collaboration. Using the same collected demonstrations, we separately perform supervised fine-tuning (SFT) of OpenVLA-OFT, $\pi_0$, and Evo-World before evaluating their task performance.

% Prefer the page top; appendix float-only pages are also top-aligned.

\end{document}